%% file: pami.tex
\documentclass[10pt,journal,compsoc]{IEEEtran}
\ifCLASSOPTIONcompsoc
  \usepackage[nocompress]{cite}
\else
  \usepackage{cite}
\fi

\input{math_commands.tex}

\usepackage[T1]{fontenc}

\usepackage{xcolor,soul,framed} 

\colorlet{shadecolor}{yellow}
\usepackage[pdftex]{graphicx}
\graphicspath{{../pdf/}{../jpeg/}}
\DeclareGraphicsExtensions{.pdf,.jpeg,.png}

\usepackage{array}
\usepackage{mdwmath}
\usepackage{url}
\usepackage{chngcntr}
\usepackage{graphicx}
\usepackage{wasysym}
\usepackage{multirow}
\usepackage{cite}
\usepackage{amsmath}
\usepackage{amssymb}
\usepackage{booktabs}
\usepackage{setspace}
\usepackage{enumerate}
\usepackage{pifont}
\usepackage{ragged2e}
\usepackage{hhline}
\usepackage{multirow}
\usepackage{colortbl}
\usepackage{mdframed}
\usepackage{subcaption}
\makeatletter  
\newif\if@restonecol  
\makeatother

\usepackage[linesnumbered,ruled, vlined]{algorithm2e}
\usepackage{algpseudocode}  
\usepackage[pagebackref,breaklinks,colorlinks]{hyperref}

\begin{document}
%
\title{StructChart: On the Schema, Metric, and Augmentation for Visual Chart Understanding}

\author{Renqiu~Xia, Haoyang Peng, Hancheng Ye, Mingsheng Li, Xiangchao Yan,\\ Peng Ye, Botian Shi, Yu Qiao, Junchi Yan and Bo Zhang
\IEEEcompsocitemizethanks{
\IEEEcompsocthanksitem Corresponding authors: Bo Zhang and Junchi Yan.
\IEEEcompsocthanksitem Renqiu Xia and Junchi Yan are with MoE Key Lab of Artificial Intelligience, Department of Computer Science and Engineering, and School of Artificial Intelligience, Shanghai Jiao Tong University. E-mail: \{xiarenqiu, yanjunchi\}@sjtu.edu.cn.
\IEEEcompsocthanksitem Hancheng Ye, Xiangchao Yan, Botian Shi, Yu Qiao and Bo Zhang are with Shanghai Artificial Intelligence Laboratory, Shanghai 200232, China. E-mail: \{yehancheng, yanxiangchao, shibotian, qiaoyu, zhangbo\}@pjlab.org.cn.
\IEEEcompsocthanksitem Haoyang Peng, Mingsheng Li and Peng Ye are with the School of Information Science and Technology, Fudan University. E-mail: \{21210720208, limc22, 20110720039\}@m.fudan.edu.cn.
\IEEEcompsocthanksitem This work was partially performed when Renqiu Xia is a research intern at Shanghai Artificial Intelligence Laboratory, China.}
} 

\markboth{IEEE TRANSACTIONS ON Pattern Analysis and Machine Intelligence}%
{Xia \MakeLowercase{\textit{et al.}}: StructChart: On the Schema, Metric, and Augmentation for Visual Chart Understanding}

\newcommand{\method}{SPOT}

\input{section/0-abstract}
\maketitle

\IEEEdisplaynontitleabstractindextext

%
\IEEEpeerreviewmaketitle

\input{section/1-intro}
\input{section/2-related}
\input{section/3-method}

\input{section/4-experiment}

\input{section/5-conclusion}

\section*{Acknowledgements}
The research was supported by National Natural Science Foundation of China (Grant No. 92370201 and 62222607), the Science and Technology Commission of Shanghai Municipality (Grant No. 22DZ1100102), and by Shanghai Artificial Intelligence Laboratory and Shanghai Rising Star Program (Grant No. 23QD1401000).

\ifCLASSOPTIONcaptionsoff
  \newpage
\fi

\bibliographystyle{IEEEtran}
\bibliography{mybib}

\end{document}

%% file: math_commands.tex
\usepackage{amsmath,amsfonts,bm}

\def\eqref#1{equation~\ref{#1}}

\def\1{\bm{1}}

\DeclareMathAlphabet{\mathsfit}{\encodingdefault}{\sfdefault}{m}{sl}
\SetMathAlphabet{\mathsfit}{bold}{\encodingdefault}{\sfdefault}{bx}{n}



%% file: section/0-abstract.tex
\IEEEtitleabstractindextext{
\begin{abstract}
\justifying{
Charts are common in literature across various scientific fields, conveying rich information easily accessible to readers. Current chart-related tasks focus on either chart perception that extracts information from the visual charts, or chart reasoning given the extracted data, \textit{e.g.} in a tabular form. In this paper, we introduce \textbf{StructChart}, a novel framework that leverages \textbf{S}tructured \textbf{T}riplet \textbf{R}epresentations (STR) to achieve a unified and label-efficient approach to chart perception and reasoning tasks, which is generally applicable to different downstream tasks, beyond the question-answering task as specifically studied in peer works. Specifically, StructChart first reformulates the chart data from the tubular form (linearized CSV) to STR, which can friendlily reduce the task gap between chart perception and reasoning.
We then propose a \textbf{S}tructuring \textbf{C}hart-oriented \textbf{R}epresentation \textbf{M}etric (SCRM) to quantitatively evaluate the chart perception task performance. To augment the training, we further explore the potential of Large Language Models (LLMs) to enhance the diversity in both chart visual style and statistical information. Extensive experiments on various chart-related tasks demonstrate the effectiveness and potential of a unified chart perception-reasoning paradigm to push the frontier of chart understanding. All codes, models and SimChart9K data are public at} \textcolor{teal}{\url{https://github.com/UniModal4Reasoning/ChartVLM}}
\end{abstract}

\begin{IEEEkeywords}
Chart Perception, Chart Understanding, Simulation-to-Real
\end{IEEEkeywords}
}

%% file: section/1-intro.tex
\section{Introduction}
\label{sec_introd}
\IEEEPARstart{C}{harts} are effecitive tools for information visualizing, and automatically extracting the underlying information from the visual charts has become a trending topic in both machine learning~\cite{Masry2022ChartQAAB, nam2023stunt} and vision communities~\cite{Luo2021ChartOCRDE, Obeid2020CharttoTextGN, Rane2021ChartReaderAP, chen2024far}.

\begin{figure}[tb!]
\centering
\includegraphics[width=0.88\linewidth]{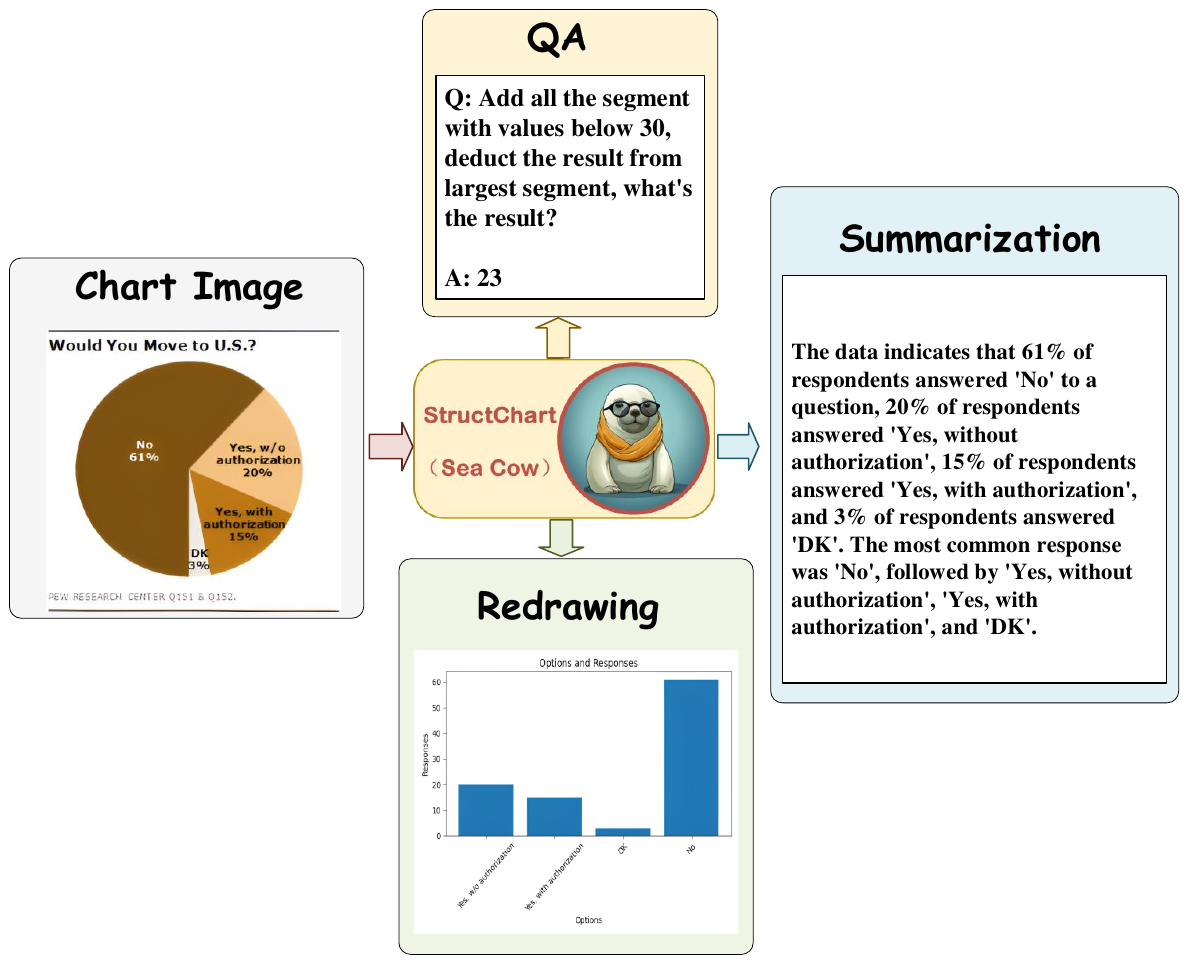}
\vspace{-5pt}
\caption{Visualization results using the proposed StructChart on different chart-related reasoning tasks including Question Answering (QA), Summarization, and Redrawing.}
\label{fig:multi_task_1}
\end{figure}

Visual Chart Understanding (CU) aims to extract the information (usually statistical data) inside a visual chart and further understanding can be made via  chart question answering or chart redrawing etc. This spans applications including medical tabular analysis~\cite{ulmer2020trust, xia2024docgenome}, chart Optical Character Recognition (OCR)~\cite{Luo2021ChartOCRDE, hegselmann2023tabllm, Obeid2020CharttoTextGN, Masry2022ChartQAAB}, and knowledge data extraction for Large Language Models (LLMs)~\cite{Brown2020gpt, chung2022scaling, he2022masked}. In general, chart-related works can be categorized into two classes (1) Chart Perception (CP) class that focuses on recognizing valuable information from a chart, converting the chart from a visual-level image to text-level representation, and (2) Chart Reasoning (CR) class that aims to understand the chart information by a tabular form. Although these works~\cite{Masry2022ChartQAAB,raffel2020exploring,Luo2021ChartOCRDE,chung2022scaling, xia2024chartx} are inspiring and have achieved promising performance gains on both tasks, joint perception-reasoning is still under-explored and challenged by the following aspects.

\input{tables/1-related-work}

\textbf{(1) Gap from Perception to Reasoning Task:} Perception task tries to extract as accurate chart information as possible, ignoring the subtle relations between data columns and rows. However, the reasoning task considers the complicated data relations to output the answer or summarize the chart information, especially for charts with both numerical and textual information.

\textbf{(2) Incomplete Metric Evaluation:} There lacks a comprehensive metric to evaluate chart perception performance from the perspective of structured information extraction with data relations. Besides, the existing metric~\cite{Masry2022ChartQAAB} only covers a single chart type such as bar~\cite{Choi2019VisualizingFT}, pie~\cite{Liu2019DataEF}, and line~\cite{Luo2021ChartOCRDE}, which can hardly generalize to other chart domains when scaling up the chart quantity. 

\textbf{(3) Expensive Chart Corpus and Annotations:} Chart data acquisition and its annotations from different fields are labor-intensive, time-consuming, and highly dependent on professionals from different fields~\cite{ulmer2020trust}.

To tackle the above challenges, we propose StructChart, a unified and label-efficient learning paradigm for joint perception and reasoning. Firstly, to \textbf{alleviate the task gap}, the image-encoder and text-decoder are devised to facilitate the representation transformation from chart images to text format. Specifically, we further develop a tailored tokenization technique that reformulates the chart from the commonly-used LCT format to a well-designed Structured Triplet Representations (STR) format, instead of the commonly-used Linearized Comma-Separated Values Tokens (LCT)~\cite{Masry2022ChartQAAB, Liu2022MatChaEV, Liu2022DePlotOV, Han2023ChartLlamaAM} which could ignore the entity relation within the chart.  Secondly, to \textbf{unify the metric evaluation}, we develop a Structuring Chart-oriented representation Metric (SCRM) based on the proposed STR, which evaluates the chart perception ability from the STR (structured information description), enabling the perception evaluation process for different types of chart data. Finally, to \textbf{expand the chart data}, we propose an LLM-based self-inspection data production scheme that generates more chart data with different domain distributions by statistical data query and drawing code generation leveraging the LLMs. We found that the chart perception and reasoning ability can be enhanced by the LLMs-based simulation method.


Experiments are conducted on both chart perception and reasoning tasks, \textit{e.g.}, chart perception, chart question answering, chart summarization, and chart redrawing. Besides, we produce a synthetic chart dataset termed SimChart9K, which significantly boosts the chart perception performance under the few-shot condition. Overall, experimental results verify that StructChart can achieve superior chart perception performance and unify the chart understanding.

The higlights of this paper can be summarized as follows:
\begin{itemize}
    \item To the best knowledge, we are the first to propose Structured Triplet Representations (STR), a groundbreaking data schema tailored for chart-related tasks. STR encapsulates the essential elements of charts into a coherent data format, enabling a more structured and interpretable representation of visual data.
    \item Leveraging the STR data schema, we introduce the Structuring Chart-oriented Representation Metric (SCRM), which is meticulously crafted to be universally applicable and adaptable to various chart-related perception tasks, providing a standardized method for evaluating and comparing chart representation effectiveness.
    \item To augment chart perception and reasoning capabilities, we produce SimChart9K, a scalable dataset generated by an LLM-based data production scheme. We observe a consistent improvement in the chart perception as more simulated charts are used for pre-training.
    \item Utilizing our proposed STR data schema and the augmented SimChart9K dataset, StructChart has achieved superior performance on QA tasks, outperforming state-of-the-art models such as DePlot~\cite{Liu2022DePlotOV}, Matcha~\cite{Liu2022MatChaEV}, ChartLlama~\cite{Han2023ChartLlamaAM} and GPT-4V~\cite{openai2023gpt4v}.
\end{itemize}

The remainder of this paper is organized as follows: Section~\ref{sec:related} briefly reviews the works on vision-language pre-trained models and chart-related models for chart perception, chart reasoning, and chart understanding. In Section~\ref{sec:method}, a Structured Triplet Representations (STR) is first proposed, and then the Structuring Chart-oriented Representation Metric (SCRM) and LLM-based self-inspection data Production Scheme is presented and its application for chart tasks is shown in details. In Section~\ref{sec:experiment}, the proposed method is evaluated on ChartQA, PlotQA, and Chart2Text benchmarks, and the experimental analyses of STR are conducted. Finally, concluding remarks are presented in Section~\ref{sec:conclusion}.

%% file: tables/1-related-work.tex
\begin{table*}[tb!]
\centering
\caption{ Comparisons of different research works on chart data, where CP, CR, and CU represent the chart perception, chart reasoning, and chart understanding works, respectively. S. and R. denote the summarization and redrawing tasks, respectively.}
\small
\scalebox{0.95}{
\begin{tabular}{c|l|ccc|c|c|cc|ccc}
\toprule
&\multirow{2}{*}{Methods} & \multicolumn{3}{c|}{Chart Types} & \multirow{2}{*}{Perception} & Reasoning / & \multicolumn{2}{c|}{Perception}  & \multicolumn{3}{c}{Downstream Tasks}         \\ 
\cline{3-5} \cline{8-12}
&  & Line & Bar & Pie& & Understanding & \multicolumn{1}{c|}{Format} & Metric & QA & S. & R.  \\ \hline
\multirow{5}{*}{\rotatebox[origin=b]{90}{\textbf{CP}}} & ReVision~\cite{Savva2011ReVisionAC}& & \checkmark &\checkmark &\checkmark & \multirow{5}{*}{\fontsize{30.0pt}{\baselineskip}\selectfont $\ \backslash$ }& \multicolumn{1}{c|}{\multirow{5}{*}{JSON}}& & \multicolumn{3}{c}{\multirow{5}{*}{\fontsize{35.0pt}{\baselineskip}\selectfont $\ \backslash$ }}  \\
& ChartReader~\cite{Rane2021ChartReaderAP}  & & \checkmark &\multicolumn{1}{c|}{} & \multicolumn{1}{c|}{\checkmark} & \multicolumn{1}{c|}{} & \multicolumn{1}{c|}{} & \multicolumn{1}{c|}{Component}  & \multicolumn{1}{l}{} & \multicolumn{1}{l}{} & \multicolumn{1}{l}{} \\
& \cite{Liu2019DataEF} & & \checkmark &\multicolumn{1}{c|}{} & \multicolumn{1}{c|}{\checkmark} &\multicolumn{1}{c|}{}  &\multicolumn{1}{c|}{}  & \multicolumn{1}{c|}{-level} & \multicolumn{1}{l}{} & \multicolumn{1}{l}{} & \multicolumn{1}{l}{} \\
& \cite{Choi2019VisualizingFT} & & & \multicolumn{1}{c|}{\checkmark} & \multicolumn{1}{c|}{\checkmark}  &\multicolumn{1}{c|}{} &\multicolumn{1}{c|}{}  & \multicolumn{1}{c|}{} & \multicolumn{1}{l}{} & \multicolumn{1}{l}{} & \multicolumn{1}{l}{} \\
\cline{9-9}
& ChartOCR~\cite{Luo2021ChartOCRDE}   & \checkmark & \checkmark  & \multicolumn{1}{c|}{\checkmark}& \multicolumn{1}{c|}{\checkmark}  &\multicolumn{1}{c|}{} & \multicolumn{1}{c|}{} & \multicolumn{1}{c|}{Type-level}  & \multicolumn{1}{l}{} & \multicolumn{1}{l}{} & \multicolumn{1}{l}{} \\ \hline
\multirow{4}{*}{\rotatebox[origin=b]{90}{\textbf{CR}}}  & T5-OCR~\cite{Masry2022ChartQAAB}  & \checkmark  & \checkmark & \multicolumn{1}{c|}{\checkmark} &\multicolumn{1}{c|}{\checkmark}  & \multicolumn{1}{c|}{\checkmark} & \multicolumn{1}{c|}{\multirow{4}{*}{LCT} }  & \multirow{4}{*}{\fontsize{30.0pt}{\baselineskip}\selectfont $\backslash$} & \checkmark&  & \\
& TaPas-OCR~\cite{Masry2022ChartQAAB} & \checkmark & \checkmark   & \multicolumn{1}{c|}{\checkmark} & \multicolumn{1}{c|}{\checkmark}  &\multicolumn{1}{c|}{\checkmark}  &\multicolumn{1}{c|}{}  & \multicolumn{1}{c|}{}  & \checkmark  &  &   \\
& VL-T5-OCR~\cite{Masry2022ChartQAAB}   & \checkmark  & \checkmark  & \multicolumn{1}{c|}{\checkmark}& \multicolumn{1}{c|}{\checkmark} &\multicolumn{1}{c|}{\checkmark} &\multicolumn{1}{c|}{}  & \multicolumn{1}{c|}{} & \checkmark & & \\
& VisionTaPas-OCR~\cite{Masry2022ChartQAAB} & \checkmark & \checkmark & \multicolumn{1}{c|}{\checkmark} & \multicolumn{1}{c|}{\checkmark} &\multicolumn{1}{c|}{\checkmark} &\multicolumn{1}{c|}{}  & \multicolumn{1}{c|}{}  & \checkmark & &  \\ \hline
\multirow{3}{*}{\rotatebox[origin=b]{90}{\textbf{CU}}} & Matcha~\cite{Liu2022MatChaEV}  & \checkmark  & \checkmark & \multicolumn{1}{c|}{\checkmark} & \multicolumn{1}{c|}{\textbf{--}}  & \multicolumn{1}{c|}{\checkmark}   &\multicolumn{1}{c|}{\textbf{--}}   & \multicolumn{1}{c|}{\textbf{--}}  & \checkmark  & \checkmark  & \\
& Deplot~\cite{Liu2022DePlotOV} & \checkmark   & \checkmark & \multicolumn{1}{c|}{\checkmark} & \multicolumn{1}{c|}{\checkmark}&\multicolumn{1}{c|}{\checkmark} & \multicolumn{1}{c|}{LCT}  &\multicolumn{1}{c|}{\textbf{--}}   & \checkmark & \checkmark  & \\
& ChartLlama~\cite{Han2023ChartLlamaAM} & \checkmark   & \checkmark & \multicolumn{1}{c|}{\checkmark} & \multicolumn{1}{c|}{\checkmark}&\multicolumn{1}{c|}{\checkmark} & \multicolumn{1}{c|}{LCT}  &\multicolumn{1}{c|}{\textbf{--}}   & \checkmark & \checkmark  &\checkmark \\
\cline{2-12} 
& StructChart (Ours)  & \checkmark  & \checkmark  & \multicolumn{1}{c|}{\checkmark} & \multicolumn{1}{c|}{\checkmark}&\multicolumn{1}{c|}{\checkmark}  & \multicolumn{1}{c|}{STR}  & \multicolumn{1}{c|}{SCRM} & \checkmark  & \checkmark & \checkmark \\ 
\bottomrule
\end{tabular}
}
\label{tab:rw_compare}
\end{table*}

%% file: section/2-related.tex
\input{tables/schema_comp} 

\section{Related Works}
\label{sec:related}
In this section, we provide an extensive survey of the current state of vision language pre-trained models and chart-related models, categorizing these models according to their architectural and functional attributes. We delve into the details of their design and operation, highlighting their relevance to our work.

\subsection{Vision Language Pre-trained Models}
According to the way of aggregating information from different modalities, Vision Language Pre-trained Models (VLPMs)~\cite{Su2019VLBERTPO,Li2020UNIMOTU,Huang2021SeeingOO,Cho2021UnifyingVT,zhang2023uni3d,Cho2021UnifyingVT,Jia2021ScalingUV,Li2021SupervisionEE} can be categorized into \textbf{fusion-encoder based}, \textbf{dual-encoder based}, and \textbf{combination based} models.

\textbf{Fusion-encoder based} VLPMs take text embeddings and image features
as input and involve several fusion approaches to model Vision Languate (VL) interaction. VL-BERT~\cite{Su2019VLBERTPO}, UNIMO~\cite{Li2020UNIMOTU}, SOHO~\cite{Huang2021SeeingOO}, VL-T5~\cite{Cho2021UnifyingVT}, SimVLM~\cite{Wang2021SimVLMSV}, and ViLT~\cite{Kim2021ViLTVT} assume that the potential correlation and alignment between modalities can be learned by a single transformer encoder. Thus, the text embeddings and image features are concatenated with additional embeddings that indicate position and modalities, and fed into a transformer-based encoder. 
Further, ViLBERT~\cite{Lu2019ViLBERTPT}, Visual Parsing~\cite{Xue2021ProbingIV}, ALBEF~\cite{Li2021AlignBF}, and WenLan~\cite{Huo2021WenLanBV} adopt a cross-attention mechanism to model the interaction between VL modalities, where the query vectors originate from one modality and the key and value vectors from the other. Typical pre-training tasks for fusion-encoder based VLPMs include: masked language/vision modeling, image-text matching, masked region classification, masked region feature regression, visual grounding, visual question answering, and grounded captioning. Thus, fusion-encoder based VLPMs can be effective in VL understanding downstream tasks.

\textbf{Dual-encoder based} VLPMs utilize two individual single-modal encoders to encode each modality separately, then convert the image and text embeddings into the same semantic space to calculate the VL similarity scores. 
CLIP~\cite{Cho2021UnifyingVT}, ALIGN~\cite{Jia2021ScalingUV} and DeCLIP~\cite{Li2021SupervisionEE} leverage large-scale image-text pairs to learn transferable visual representations for retrieval tasks and exhibit surprising zero-shot transfer to image classification tasks.

\textbf{Combination based} VLPMs combine the benefits of fusion-encoder based and dual-encoder based architectures. FLAVA~\cite{Singh2021FLAVAAF} firstly utilizes a dual-encoder to acquire single-modal representations. Then, the single-modal embeddings are processed by a fusion-encoder to obtain cross-modal representations. Vlmo~\cite{Wang2021VLMoUV} introduces a novel approach called Mixture-of-Modality Expert (MoME), combining a dual-encoder and a fusion-encoder into one unified framework, which can be fine-tuned on both VL understanding and image-text retrieval tasks.

\subsection{Synthetic Data for Augmentation}
Data has consistently been a pivotal element in the achievements of generative models. The latest advancements in these models are mainly due to the easy access to vast, diverse, and high-quality datasets needed for training. To delve into the role of synthetic data, we willwe will summarize the related work that synthesize \textbf{Language Modal} and \textbf{Multimodal} data.

\textbf{Language Modal.} Synthetic data is increasingly seen as a powerful tool for creating large and top-notch datasets. Researchers have explored various methods, from using differential privacy to building instruction-tuning frameworks, all to enhance the synthetic data’s quality, diversity, and usefulness~\cite{Lou2023MUFFINCM,Wei2023MagicoderSC}. Instruction Backtranslation\cite{Li2023SelfAlignmentWI} has developed a top instruction-following language model by automatically tagging human-written text with relevant instructions, showcasing a highly effective self-alignment technique. MAmmoTH~\cite{Yue2023MAmmoTHBM} ultilizes Self-Instruct to generate questions and chains of thought (CoT) pairs, and further generates Program of Thought (PoT), making it easier to solve mathematical tasks.

\begin{figure*}[tb!]
\centering
\includegraphics[width=0.88\linewidth]{images/StructChart_framework.pdf}
\caption{StructChart overview: 1) LLM-based production scheme for providing more chart data; 2) CIE training for chart perception; 3) Representation transformation for bridging the task gap; 4) Downstream reasoning tasks, including Question Answering, Summarization and Redrawing.}
\label{fig:framework}
\end{figure*}

\textbf{Multimodal.} ShareGPT4~\cite{Chen2023ShareGPT4VIL} uses GPT-4V~\cite{openai2023gpt4v} to generate high-quality image captions, achieving alignment and delivering cutting-edge results on models like LLaVA. StableLLaVA~\cite{Li2023StableLLaVAEV} synchronously generates images alongside dialogue,which employs ChatGPT to generate creative prompts for image generation, and then utilizes Stable Diffusion to synthesize images from these prompts.  AnyGPT~\cite{Zhan2024AnyGPTUM} is built through a two-phase strategy that involves creating text-based dialogues infused with multimodal components derived from various meta-topics, followed by the transformation of text into multimodal content. ComVint~\cite{Du2023WhatMF} is a synthetic visual reasoning instruction dataset consisting of 32K examplesemploys a pipeline consisting of synthesis, complication, and reformulation when provided with an image that includes available annotations.

\subsection{Chart-related Models}
We categorize chart-related models into three types based on the task: Chart Perception, Chart Reasoning, and Chart Understanding, among which structchart is focused on Chart Understanding (CU). The differences between StructChart and other chart-related works are shown in Table~\ref{tab:rw_compare}. We also compare our proposed STR with other chart data schemas in Table~\ref{tab:comparison}.

\textbf{Chart Perception} refers to obtaining the numerical and textual values (often in the tabular) from the charts. Earlier works are based on manually designed feature extraction (\textit{e.g.} color continuous searching and edge extraction). ReVision~\cite{Savva2011ReVisionAC} employs a set of hand-crafted features and rules (color continuous searching) to extract salient marks for chart values inferring. 
ChartReader~\cite{Rane2021ChartReaderAP} takes a combined approach, using rule or heuristic-based edge extraction supported by OCR for text elements. To avoid brittle hand-crafted features, some works leverage object detection methods and OCR to extract chart value for chart perception task. 
\cite{Choi2019VisualizingFT} adopted the idea of general object detection to detect the bar components by treating each bar as an object. For Pie Chart, \cite{Liu2019DataEF} proposes to use the recurrent network and feature rotation mechanism to extract the data. 
ChartOCR~\cite{Luo2021ChartOCRDE} employs a modified version of CornerNet~\cite{Law2018CornerNetDO} backbone for keypoint detection to reconstruct the chart components (\textit{e.g.} bars and sectors), and OCR for component value.

\textbf{Chart Reasoning} seeks for chart image information to execute logical or mathematical reasoning, where Question Answering (QA) is a typical reasoning task for chart data. 
Prior works pass extracted table data and questions to TableQA models.
T5-OCR~\cite{Masry2022ChartQAAB} and TaPas-OCR~\cite{Masry2022ChartQAAB} employ ChartOCR~\cite{Luo2021ChartOCRDE} to extract table data from chart images, and conduct QA task powered by T5~\cite{Nan2021FeTaQAFT} and TaPas~\cite{Herzig2020TaPasWS} respectively. Some works adopt the two-stage reasoning pipeline, improving the TableQA models~\cite{pasupat2015compositional}. 
VL-T5-OCR~\cite{Masry2022ChartQAAB} and VisionTaPas-OCR~\cite{Masry2022ChartQAAB} extend cross-modality encoder in T5~\cite{raffel2020exploring} and TaPas~\cite{Herzig2020TaPasWS} with chart image features. Besides, Pix2Struct~\cite{Lee2022Pix2StructSP} uses the screenshot parsing input to perform self-supervised pre-training from abundant website data. ChartT5~\cite{Zhou2023EnhancedCU} utilizes cross-modal pre-training on chart-table pairs.

\textbf{Chart Understanding} is at a wider level than chart reasoning (at least by the scope of this paper), covering more open-ended and high-level tasks. Besides question answering task, chart understanding contains a wider variety of generative tasks, such as chart summarization, chart redrawing, \textit{etc}. Matcha~\cite{Liu2022MatChaEV} and Deplot~\cite{Liu2022DePlotOV} are the pioneering attempts for chart understanding, with both carrying out the QA and summarization tasks. Matcha~\cite{Liu2022MatChaEV} pre-trains a Pix2Struct~\cite{Lee2022Pix2StructSP} with chart derendering and math reasoning tasks, while Deplot~\cite{Liu2022DePlotOV} harnesses Vision Language Pre-trained Model (VLPM) to extract chart information, and subsequently employs LLMs to conduct inference for the QA and summarization. ChartLlama~\cite{Han2023ChartLlamaAM} fine-tunes LLaVA~\cite{Liu2023ImprovedBW} on private data for CU tasks.

%% file: tables/schema_comp.tex
\newcolumntype{P}[1]{>{\centering\arraybackslash}p{#1}}

\begin{table*}[tb!]
\centering
\footnotesize
\caption{Comparison of data schemas for chart perception tasks, including Comma Separated Values (CSV), Markdown (MD), JavaScript Object Notation (JSON) and our proposed Structured Triplet Representation (STR).}
\label{tab:comparison}
\begin{tabular}
{p{0.115\linewidth}|p{0.195\linewidth}p{0.195\linewidth}p{0.18\linewidth}p{0.22\linewidth}}
\toprule
& \textbf{CSV} & \textbf{MD} & \textbf{JSON} & \textbf{STR} \\
\toprule
\textbf{Evaluation\,\,\,\, Suitability}     & Not optimized for perceptual task evaluation. & Similar to CSV with limited suitability. & Good for evaluation but requires custom parsing. & High suitability for SCRM-based structured evaluation. \\
\hline
\textbf{Information Proximity}       & Information can be distant, reducing proximity. & The same as CSV. & Proximity depends on the document’s structure. & High proximity due to grouping of related entities. \\
\hline
\textbf{Scalability}       & Limited scalability with complex relations. & Limited scalability. & High scalability by nested structure.   & High scalability by tuple structure for entity relationships.  \\
\hline
\textbf{Syntax Complexity}   & Simple but less structured for complex data. & Simple but less structured for complex data. & Syntax can be complex due to nested structures.    & Simple and intuitive syntax for chart perception tasks.\\
\hline
\textbf{Human Readability}    & Limited readability for 
complex or long-sequence data. & Limited readability for complex or long-sequence data. & Readable with proper formatting but can be dense.  & High readability for perception tasks of complex data.\\

\bottomrule
\end{tabular}
\end{table*}

%% file: section/3-method.tex
\section{The Proposed Method}
\label{sec:method}

StructChart includes four key components: \textbf{\textit{(1) Transformer-based Chart-oriented Information Extractor (CIE)}}. It incorporates an image-encoder and text-decoder to facilitate the transformation from chart images to CSV-format texts. \textbf{\textit{(2) Structured Triplet Representations (STR).}} The extracted intermediate CSV text is structured into a triplet form to elucidate the intricate position relationship between the header and index.
\textbf{\textit{(3) Structuring Chart-oriented Representation Metric
(SCRM).}} A metric is designed to evaluate the quality of converted triplets, which facilitates the subsequent reasoning. \textbf{\textit{(4) LLM-based Self-inspection Data Production Scheme.}} A novel paradigm of chart data simulation is developed to enhance zero/ few-shot perception and reasoning ability, continuously improving performance by scaling up the simulated charts. Fig.~\ref{fig:framework} shows the overall paradigm.

\subsection{Overall Design for Two-stage StructChart}
Unlike end-to-end multi-modal tasks~\cite{Masry2022ChartQAAB, Obeid2020CharttoTextGN}, we fulfill CU by jointly performing perception and reasoning tasks, with STR serving as a bridge between them.

\noindent\textbf{Perception Stage.}
We propose CIE that utilizes a pixel-level encoder and a text-level decoder, where the vision encoder is based on ViT~\cite{Dosovitskiy2020AnII}. Unlike conventional fixed input resolution for ViT, CIE dynamically rescales images to maintain a constant patch number, ensuring it adheres to the predefined maximum sequence length. Moreover, we add absolute positional embeddings for the input patches, allowing the perception module to process images of varying resolutions. 
At this stage, the chart at pixel level can be converted to text-level Linearized CSV Tokens (LCT).

\noindent\textbf{Reasoning Stage.}
\label{sec:structchart model}
Before performing the reasoning process, we convert LCT into the designed STR to facilitate the module's understanding of chart-oriented information (see Sec.~\ref{subbsection:reformulation} for details). This structuring process provides reasoning with a better understanding of the entity relation within a chart. Considering the difficulty in evaluating downstream tasks, the reasoning process is performed on the QA task with various LLMs in a zero-shot way.


The motivations of the designed two-stage pipeline are: 1) explicit chart image representations can enhance the interpretability of the subsequent reasoning stage, and 2) the extracted chart data (STR) can be used as the pre-training corpus for large language models and vision-language models.

\subsection{Schema: Structured Triplet Representations (STR) for Chart Understanding}
\label{subbsection:reformulation}
Visual charts often contain rich textual and numerical information. Generally, the chart information is represented by long-form texts in the form of CSV, \textit{i.e.} previously-mentioned LCT. However, the LCT format is sensitive to positional variations of entities from charts due to its linear form. 
Thus, we propose to reformulate the LCT format to effectively and robustly represent the positional relations between row and column headers of a given chart.

\noindent\textbf{Task Definition and Structuring.} 
Given a chart image, the extracted LCT can be described as:
\begin{equation}
\small
\begin{aligned}
C_{csv} := &\mathrm{none},...,\mathrm{Entity}_{c_m},...\mathrm{Entity}_{c_M} \ /\mathrm{n}\\
&\mathrm{Entity}_{r_1},...,\mathrm{Value}_{r_1}^{c_m},...,\mathrm{Value}_{r_1}^{c_M} \ /\mathrm{n}\\
&......\\
&\mathrm{Entity}_{r_n},...,\mathrm{Value}_{r_n}^{c_m},...,\mathrm{Value}_{r_n}^{c_M} \ /\mathrm{n}\\
&......\\
&\mathrm{Entity}_{r_N},...,\mathrm{Value}_{r_N}^{c_m},...,\mathrm{Value}_{r_N}^{c_M} \ /\mathrm{n} \ ,
\end{aligned}
\label{equ:r_csv}
\end{equation}
where {\small $/\mathrm{n}$} denotes line break, and {\small $\mathrm{Entity}_{r_n}$}, {\small $\mathrm{Entity}_{c_m}$} indicate {\small $n$-th} row header entity and {\small $m$-th} column header entity, respectively, {\small $M,N\in \mathbb{{N}^+}$}. {\small $\mathrm{Value}_{r_n}^{c_m}$} in Eq.~\ref{equ:r_csv} contains positional information of {\small $\mathrm{Entity}_{c_m}$} and {\small $\mathrm{Entity}_{r_n}$}. However, LCT still faces two issues: \textit{(1) The evaluation of the predicted long-form texts containing positional information for perception model selection is non-trivial.} \textit{(2) The position-sensitive property of LCT increases the inference difficulty of different downstream chart tasks.}

\begin{figure}[tb!]
\centering
\includegraphics[width=1.02\linewidth]{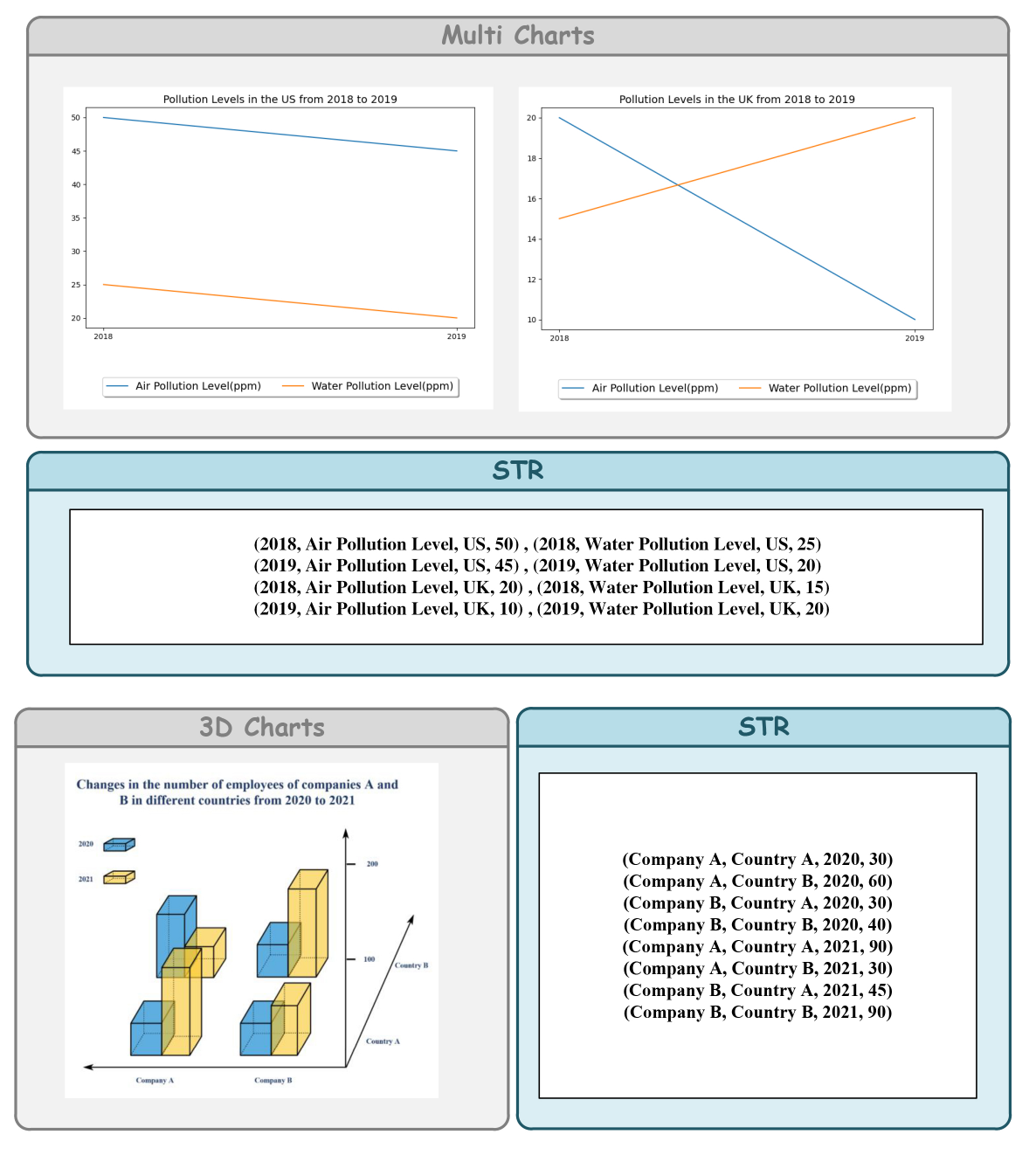}
\caption{Two examples of STR that are extended to represent high-order relations in multi-charts and high-dimensional charts.}
\label{fig:str_expansion}
\end{figure}


Since chart information has matrix-like row-column transformation invariance and transpose transformation invariance, LCT is structured into a well-designed triplet, \textit{i.e.,} STR. Given LCT tokens {\small $C_{csv}$} shown in Eq.~\ref{equ:r_csv}, STR is fomulated as Eq.~\ref{equ:r_tri}.
\begin{equation}
\small
\begin{aligned}
C_{tri} := &(\mathrm{Entity}_{r_1}, \mathrm{Entity}_{c_1},\mathrm{Value}_{r_1}^{c_1}), \\
&........, \\
&(\mathrm{Entity}_{r_n}, \mathrm{Entity}_{c_m},\mathrm{Value}_{r_n}^{c_m}),\\
&........, \\
&(\mathrm{Entity}_{r_N}, \mathrm{Entity}_{c_M},\mathrm{Value}_{r_N}^{c_M}). \\
\end{aligned}
\label{equ:r_tri}
\end{equation}

\noindent\textbf{Scalability of STR for High-order Relations.}
\label{sec:str_exten}
In a single simple chart, relations of entities are usually low-order (usually two entity variables).
For multi-charts and high-dimension charts, STR can be extended to represent high-order relations with multiple variables. We also provide visual demonstration in Fig.~\ref{fig:str_expansion} to illustrate high-order STR that represents multi-charts and high-dimensional charts.
Given the entity variables {\small$Entity_{X_i}, Entity_{Y_j}, Entity_{Z_k}, ....$}, the high-order expansion of STR can be expressed as:
\begin{equation}
\small
\begin{aligned}
C_{tri} := &(Entity_{X_1},  Entity_{Y_1}, Entity_{Z_1}, ..., Value_{1\times1\times1}), \\
&........, \\
&(Entity_{X_i},  Entity_{Y_j}, Entity_{Z_k},..., Value_{i\times j\times k}),\\
&........, \\
&(Entity_{X_I},  Entity_{Y_J}, Entity_{Z_K},..., Value_{I\times J\times K}).\\
\end{aligned}
\end{equation}

\subsection{Metric: Structuring Chart-oriented Representation Metric (SCRM) to Evaluate Chart Perception}
\label{sec:metric}
Furthermore, SCRM is designed to comprehensively evaluate the extracted chart information represented by STR. When comparing the predicted STR and ground-truth (GT) STR $C_{tri}$, we treat {\small $\mathrm{Entity}_{r_n}, \mathrm{Entity}_{c_m}$} as strings and {\small$ \mathrm{Value}_{r_n}^{c_m}$} as floats, respectively. 

\textbf{1) Image-level}. Suppose there are totally {\small $\mathbf{P}$} triplets in the predicted STR and {\small $\mathbf{Q}$} triplets in GT STR, SCRM per image is computed as follows:

\begin{enumerate}[-]
\item For {\small $\mathrm{Entity}$}, we obtain the edit distance of the $p$-th prediction string and the $q$-th GT string: 
\begin{equation}
\label{edit_dis}
\small
\begin{aligned}
J(p,q)=1-\frac{| \mathrm{Entity}_{pred}^p \cap \mathrm{Entity}_{GT}^q \mid}{\mid \mathrm{Entity}_{pred}^p \cup \mathrm{Entity}_{GT}^q \mid} \ .    
\end{aligned}
\end{equation}

\item For {\small $\mathrm{Value}$}, we calculate the relative error between the $p$-th prediction value and the $q$-th GT value: 
\begin{equation}
\label{num_tol}
\small
e(p,q)=\left|\frac{\mathrm{Value}_{pred}^{p}-\mathrm{Value}_{GT}^{q}}{\mathrm{Value}_{GT}^q}\right| \ .
\end{equation}

\item To achieve a comprehensive evaluation, we design three levels of tolerance for fine-grained judgment, aiming to measure the similarity between the predicted triplets and GT triplets, by calculating the Intersection over Union {\small $IoU|_{tol}$}, under the given tolerance level {\small $tol$} as follows:
\begin{equation}
\small
l(p,q)|_{tol}=\left\{\begin{matrix}
 1, & J(p,q) \le J_{thr}|_{tol} \wedge e(p,q)\le e_{thr}|_{tol} \\
 0, & else
\end{matrix}\right.,
\end{equation}
\begin{equation}
\small
\begin{aligned}
tol:= &\{ strict, slight, high\}, \\
strict:= &\left\{J_{thr}|_{tol}=0   \wedge   e_{thr}|_{tol}=0\right\}, \\
slight:= &\left\{J_{thr}|_{tol}=2  \wedge  e_{thr}|_{tol}=0.05\right\}, \\  
high:= &\left\{J_{thr}|_{tol}=5  \wedge e_{thr}|_{tol}=0.1\right\}, \\
\end{aligned}
\end{equation}

\begin{equation}                                 
\label{eq:iou}
\small
IoU|_{tol}=\frac{\sum_{q=1}^{\mathbf{Q}}\sum_{p=1}^{\mathbf{P}}l(p,q)|_{tol}}{\mathbf{P}+\mathbf{Q}-\sum_{q=1}^{\mathbf{Q}}\sum_{p=1}^{\mathbf{P}}l(p,q)|_{tol}}.
\end{equation}
\end{enumerate}

\textbf{2) Dataset-level}. Given the dataset with {\small $\mathbf{L}$} chart images ({\small $\mathbf{L}\in \mathbb{{N}^+}$}), the Intersection over Union of the {\small $i$-th} image can be denoted as {\small $IoU(i)$}. Besides, given a preset similarity threshold {\small $IoU_{thr}$}, the corresponding discriminant function towards the positive and negative images can be written as:
\begin{equation}
\small
d(i)|_{IoU_{thr},tol}=\left\{\begin{matrix}
  1,& if: IoU(i)|_{tol} \ge IoU_{thr} \\
  0,& else
\end{matrix}\right..
\end{equation}

When the preset similarity threshold {\small $IoU_{thr}$} becomes a variable (denoted as $t$), it changes to:
\begin{equation}
\small
d(i, t)|_{tol}=\left\{\begin{matrix}
  1,& if: IoU(i)|_{tol} \ge t \\
  0,& else
\end{matrix}\right..
\end{equation}

SCRM consists of two indicators ($Precision$ with a fixed similarity threshold and $mPrecision$ with a varying one in the range {\small $(0.5:0.05:0.95)$}):
\begin{equation}
\begin{aligned}
AP|_{IoU_{thr},tol}=&\frac{\sum_{i=1}^{L}d(i)|_{IoU_{thr},tol}}{L},
\\
\label{eqa:mprecison}
mAP|_{tol}=&\frac{\sum_{t=10}^{19}\sum_{i=1}^{L}d(i,0.05t)|_{tol}}{10L} .
\end{aligned}
\end{equation}

\subsection{Augmentation: Simulating Charts with Enhanced Diversity for Pretraining-Finetuning}
\label{sec:simulate}
Considering the difficulty of chart data acquisition and labeling cost, we introduce an LLM-based text-to-chart level data production scheme, namely PlotAgent, involving: \textit{(1) statistical data query to ensure the data diversity}, and \textit{(2) drawing code generation to ensure the style diversity.} 

\noindent\textbf{Data Query.}
Denote {\small $\mathbb{D}_{ori}=\{\mathbf{I}_{ori},\mathbf{T}_{ori}\}$} as a chart dataset, where {\small $\mathbf{I}_{ori}=\{I_{ori}^1,I_{ori}^2,...,I_{ori}^n,...,I_{ori}^N\}$ } are images and {\small $\mathbf{T}_{ori}=\{T_{ori}^1,T_{ori}^2,...,T_{ori}^n,...,T_{ori}^N\}$ } are the corresponding labeled texts in CSV format. At this stage, we aim to imitate labeled CSV texts {\small $\mathbf{T}_{ori}$ } in original chart dataset $\mathbb{D}_{ori}$ through the proposed PlotAgent to generate simulated CSV labels {\small $\mathbf{T}_{sim}$}. In detail, We employ the few-shot prompting, leveraging generative model~\cite{OpenAI2023GPT4TR} to complete the imitation task. For diversity and effectiveness of the imitation data, we impose three restrictions on the demonstration and instruction: \textit{(1) The simulated content {\small $T_{sim}^n$ } must be in CSV format}, \textit{(2) The scale of {\small $T_{sim}^n$ } can be altered compared with {\small $T_{ori}^n$ }, including the row number}, and \textit{(3) The combination of text in {\small $T_{sim}^n$ } must be reasonable, even though it may be highly irrelevant to {\small $T_{ori}^n$.}}

\noindent\textbf{Code Generation.}
The simulated CSV labels {\small $\mathbf{T}_{sim}=\{T_{sim}^1,T_{sim}^2,...,T_{sim}^n,...,T_{sim}^N\}$}, are then used to create images {\small$\mathbf{I}_{sim}= \{I_{sim}^1,I_{sim}^2,...,I_{sim}^n,...,I_{sim}^N\}$}. We still employ PlotAgent to directly generate the drawing code for the chart plotting {\small $I_{sim}^n$ } based on {\small $T_{sim}^n$}. To guarantee the diversity of chart distribution at the image level, we implement the following limitations within the instruction prompting: \textit{(1) Random sampling of a chart type that is appropriate for {\small $T_{sim}^n$}}, {\textit{comprising histograms, scatterplots, line charts, and pie charts}, \textit{(2) Random setting of drawing style}, \textit{including fonts, colors, line styles, backgrounds, \textit{etc}}, and \textit{(3) Transformation of scale with different coordinate axes.}

\begin{figure*}[tb!]
    \centering
  \subfloat[\small Real Datasets.\label{fig:tsne_real}]{%
       \includegraphics[width=0.48\linewidth]{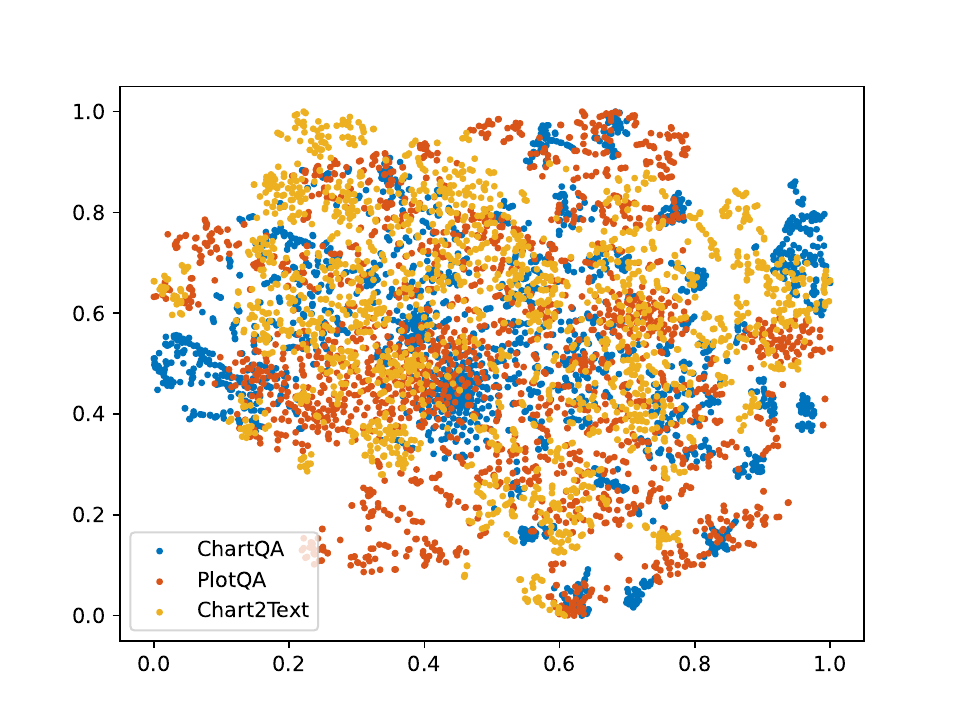}}
    \hfill
  \subfloat[\small Real \& Simulated Datasets.\label{fig:tsne_simulated}]{%
        \includegraphics[width=0.48\linewidth]{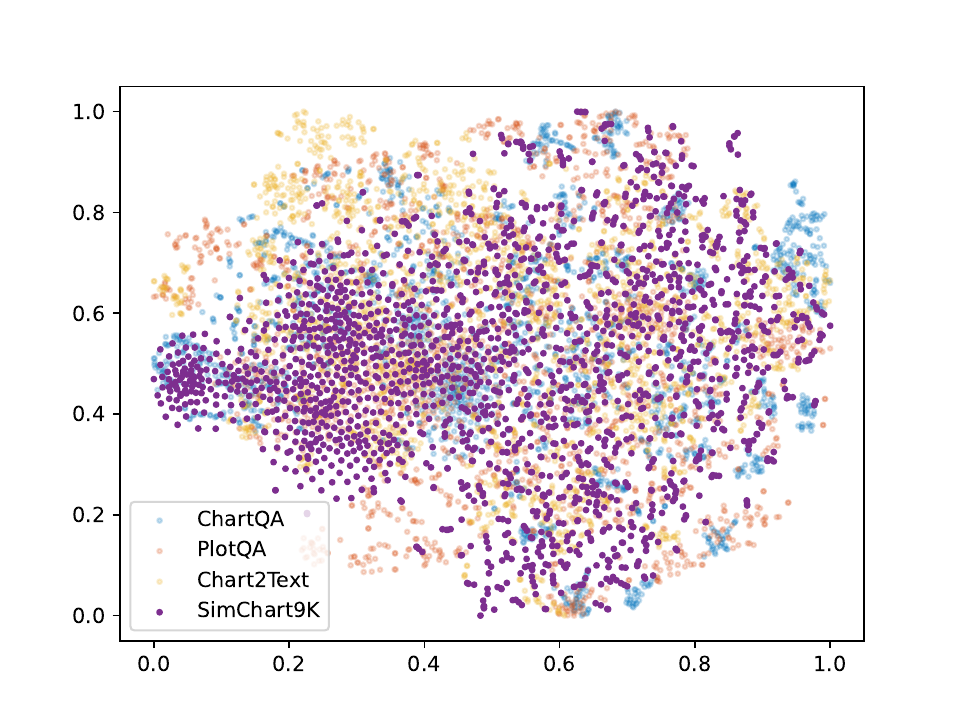}}
  \caption{Feature distributions of ChartQA, PlotQA, Chart2Text, and SimChart9K visualized by t-Distributed Stochastic Neighbor Embedding (t-SNE)~\cite{van2008visualizing}. Note that all feature embeddings of chart images are compressed to 2-dimension.}
  \label{fig:tsne} 
\end{figure*}

To guarantee that the generated drawing code can be executed correctly, we design a self-inspection mechanism to iteratively skip the non-executable code generated by our PlotAgent, until all the drawing code that matches the corresponding text can be executable. Benefiting from the text diversity during generation, we can obtain diverse statistical data and drawing codes. As a result, the proposed LLM-based text-to-chart level data production scheme can be used to simulate scalable, data-rich and style-diverse chart dataset {\small $\mathbb{D}_{sim}=\{\mathbf{I}_{sim},\mathbf{T}_{sim}\}$ } based on scale-invariant or few-shot original chart datasets {\small $\mathbb{D}_{ori}=\{\mathbf{I}_{ori},\mathbf{T}_{ori}\}$ }.

%% file: section/4-experiment.tex
\section{Experiments}
\label{sec:experiment}
\subsection{Experimental Setup}
\label{subsec:datasets}

\noindent\textbf{Datasets:} 
We evaluate StructChart on four real-world chart benchmarks and our simulated dataset with image-CSV pairs. \textbf{ChartQA}~\cite{Masry2022ChartQAAB} is a large-scale visual reasoning dataset with 20,882 charts collected from online sources, composed of an \textit{augmented} set generated synthetically and a \textit{human} set written by humans. \textbf{PlotQA}~\cite{Methani2019PlotQARO} is a synthetic dataset covering 28.9 million QA pairs across 224,377 plots. \textbf{FigureQA}~\cite{Kahou2017FigureQAAA} is a visual reasoning corpus consisting of over one million human-designed QA pairs. \textbf{Chart2Text}~\cite{Obeid2020CharttoTextGN} is for automatic summarization of statistical charts crawled from \texttt{statista.com}, yielding 8,305 charts with labels. \textbf{SimChart9K} is the proposed simulated dataset composed of 9,536 charts and corresponding CSV labels, which is generated via ChartQA using PlotAgent.

\input{tables/2-real-se}
\input{tables/3-sim-se}

\noindent\textbf{Implementation Details:} 
We design various data consolidation settings based on both real and simulated data to train StructChart. For perception, we use image-CSV pairs in ChartQA, PlotQA, and SimChart9K to train the Chart-oriented Information Extractor with 300M parameters using Cross-Entropy loss. All the labels are in LCT representation. For reasoning, we transfer the prediction (LCT) from the perception stage to STR representation and use various LLMs (i.e., Llama2, Vicuna-v1.5, Llama3.1, GPT-4, etc) without prompt engineering to perform various downstream tasks for fair comparisons with other methods. All the training experiments are conducted with 8 A100 GPUs.

\noindent\textbf{Evaluation Metric:}
We implement our proposed SCRM to evaluate the perception task (from image to CSV) and relaxed-acc~\cite{Masry2022ChartQAAB} for QA task. In the relaxed-acc evaluation metric, a 5\% error margin is allowed for numerical answers, while for string answers, an Exact Match (EM) is required.


\input{tables/4-fewshot}
\input{tables/6-chartqa}

\subsection{Chart Perception on Real-world and Simulation Data}
\label{sec:real_sim}
We first conduct experiments on real-world datasets including ChartQA, PlotQA, and Chart2Text, and further merge them for joint-dataset training to evaluate the performance scalability given more training data. From Table~\ref{tab:real}, with a comprehensive study on ChartQA, PlotQA, and Chart2Text, StructChart continuously improves the perception performance of chart data on each domain given more training samples. Moreover, Fig.~\ref{fig:tsne_real} visualizes the feature distributions,
showing notable differences in the feature distributions across distinct datasets.

We randomly divide the proposed SimChart9K dataset into four subsets with different amounts (0.1K, 1K, 6K, 9K), and train the StructChart on the mixed dataset (including ChartQA training set and SimChart with different amounts). All evaluations are conducted on the ChartQA validation set based on the proposed SCRM metrics. In Table~\ref{tab:sim}, it can be seen that the introduction of simulation dataset significantly improves the CIE performance of StructChart; that is, \textbf{the larger the simulation dataset, the greater the performance gains in CIE}. Furthermore, we also illustrate feature distributions of SimChart9K in Fig.~\ref{fig:tsne_simulated}. It can be observed from Fig.~\ref{fig:tsne_simulated} that the feature distribution of the simulated data (SimChart9K) can be better matched with that of the real data such as ChartQA, PlotQA, and Chart2Text. As a result, the simulated charts are beneficial to boost the CIE performance of the model towards the real-world charts.

Note that Matcha~\cite{Liu2022MatChaEV} and Deplot~\cite{Liu2022DePlotOV} in Chart Information Extraction (CIE) task are not reported as baselines for Chart2Text dataset in Table~\ref{tab:real}. The reason is that the charts in the Chart2Text dataset are accompanied by background noise from the web environment. As a result, Matcha tends to output HTML-formatted garbled text, while Deplot is also affected by the background noise, leading to the generation of irrelevant content. This interference from the environmental noise poses challenges for these methods, making them less suitable as direct baselines for the Chart2Text dataset. By comparison, Structchart can extract accurate information of the chart from complicated backgrounds, such as information from the website.

\subsection{Zero/Few-shot Learning on SimChart9K with Limited Real-world Chart Data}
\begin{figure}[tb!]
\centering
\includegraphics[width=0.4\textwidth]{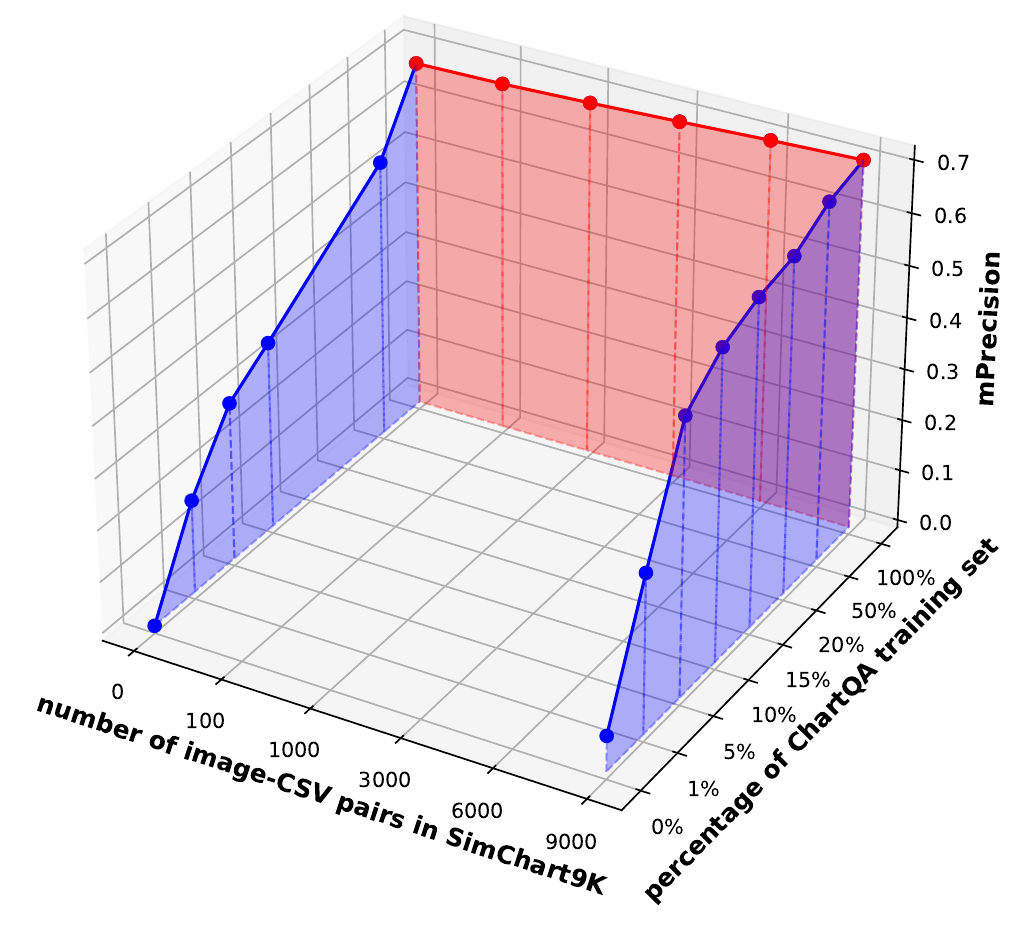}
\caption{Few-shot learning performance with simulated data and limited real data: mPrecision by various amounts of charts in SimChart9K \& ChartQA.}
\label{fig:heat}
\end{figure}

The purpose of this part is to answer two questions: 1) Can we achieve a high-performance CIE only leveraging few-shot real samples? 2) With the help of SimChart9K, how many real-world samples can obtain the CIE performance that is achieved on the full training set? To answer these questions, we split the real chart dataset ChartQA into subsets with different sizes, including subsets with 1\%, 10\%, 20\% and 50\% original real-world samples. We demonstrate zero-shot and few-shot results in Table~\ref{tab:few-shot}, obtaining the following observations: (1) When the model is trained on the real dataset ChartQA alone without any simulated samples, the CIE performance is still positively correlated with the number of real-world training samples. (2) Training only on the simulated dataset without any real samples (zero-shot training) fails to achieve a satisfactory CIE performance, due to the insufficiency in real-world charts. (3) By leveraging the proposed method to generate many simulated charts (SimChart9K), only 20\% real-world charts can basically achieve equal CIE performance under the 100\% real-world training samples. Besides, Table~\ref{tab:few-shot} indicates that the CIE performance obtained using 50\% real-world charts significantly outperforms that obtained using the full-set training examples. We further summarize the above observations in Fig.~\ref{fig:heat}. Besides, we conduct experiments on PlotQA and Chart2Text, and the findings presented in Table~\ref{tab:few-shot-rebuttal} underscore the generalization capabilities that SimChart9K provides for StructChart. Specifically, leveraging SimChart9K, only 10\% original real samples in PlotQA and 20\% in Chart2Text can achieve equivalent CIE performance under the 100\% real training samples. 

\subsection{From Perception to Reasoning: STR}
\label{sec:exp_qa}

We conduct experiments to verify the effectiveness of STR as the intermediate representations of chart for various downstream reasoning tasks. We compare STR with commonly-used LCT. Besides, we leverage multiple mainstream LLMs as the reasoning module in a zero-shot prompting way to validate the effectiveness of StructChart.

\noindent\textbf{For QA Task}, we evaluate on ChartQA~\cite{Masry2022ChartQAAB} using the Exact Match (EM) metric for the text answer, where a 5\% tolerance for the numerical answer is allowed to make a fair comparison. Table~\ref{tab:qa} shows that: (1) Our proposed two-stage pipeline (from perception to reasoning) and STR representation have been verified to be effective across the majority of LLMs, including Llama2, Vicuna-v1.5, Llama3.1 and GPT-4. (2) By comparing LCT, the proposed STR facilitates a better understanding of LLMs, yielding higher answer accuracies. In our analysis, this is mainly due to that the STR is designed using a structural description, with a row-column matching relation compared with the previous LCT format. (3) StructChart+GPT4 pipeline surpasses all baselines on the ChartQA val\&test set, especially exceeding the end-to-end commercial model GPT-4V.

\begin{figure}[htbp]
    \centering
       \includegraphics[width=0.89\linewidth]{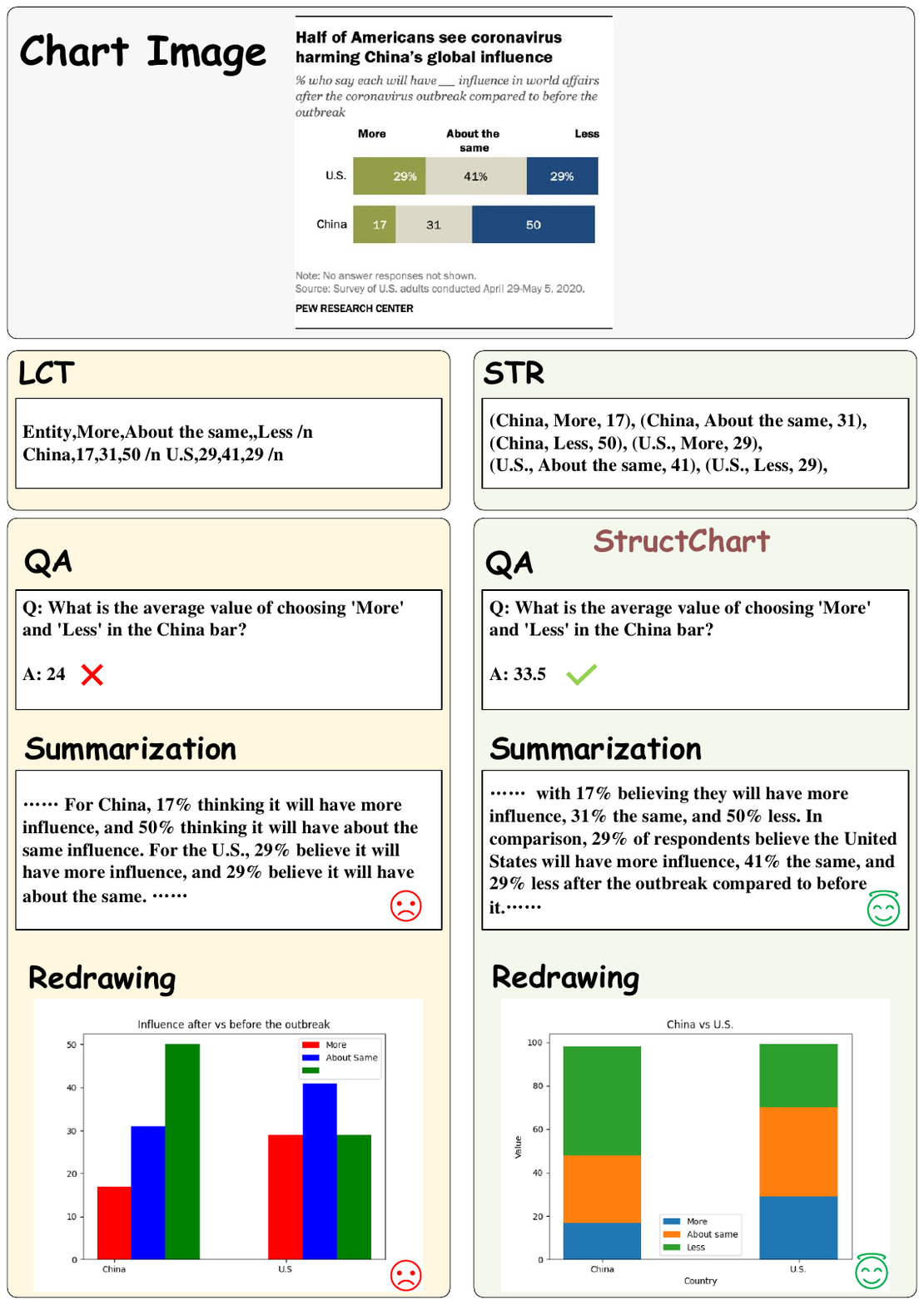}
  \caption{Visualization results using the proposed StructChart on downstream tasks with Linear CSV Tokens (LCT) v.s. Structured Triplet Representations (STR).}
  \label{fig:lct_str} 
\end{figure}

\begin{figure}[htbp]
    \centering
       \includegraphics[width=0.89\linewidth]{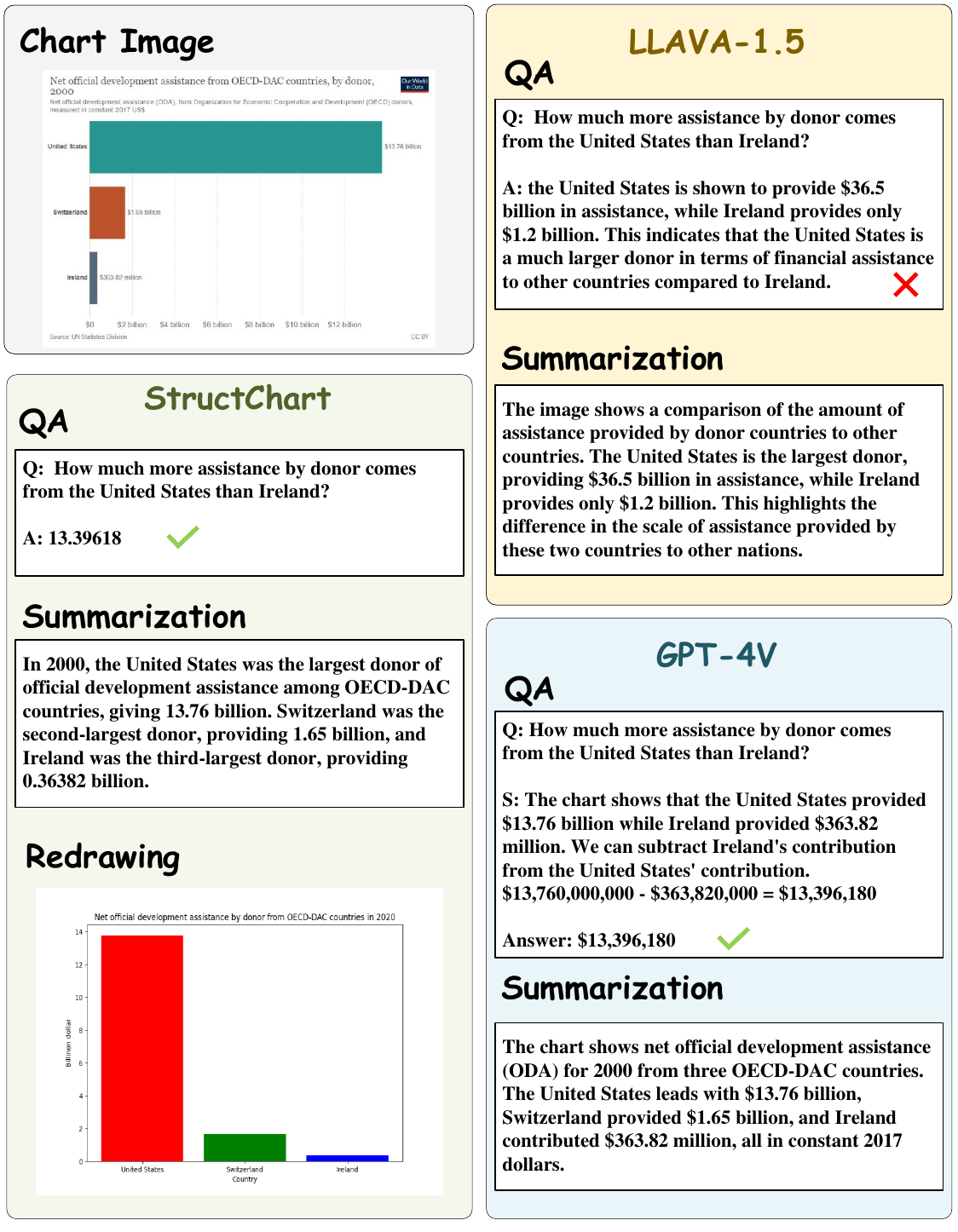}
  \caption{Visualization comparisons among StructChart, LLaVA-1.5~\cite{Liu2023VisualIT} and GPT-4V~\cite{openai2023gpt4v} on downstream tasks.}
  \label{fig:gpt4v} 
\end{figure}

\noindent\textbf{For Summarization and Redrawing Tasks}, due to the lack of public benchmark and reasonable metrics, it is difficult to provide quantitative results. Thus, we present qualitative results employing different intermediate representations (LCT and STR) in Fig.~\ref{fig:lct_str}, which further shows that the STR has a stronger ability to represent chart information and help chart understanding. We provide comparison examples of StructChart with other general VLMs (\textit{i.e.} GPT-4V~\cite{openai2023gpt4v} and LLaVA-1.5~\cite{Liu2023VisualIT}) on downstream tasks in Fig.~\ref{fig:gpt4v}. These results underscore the advantage of STR in tasks requiring detailed chart understanding, as it enables a more structured and semantically rich representation of the chart data. This aligns with the broader objective of improving downstream tasks, such as summarization and redrawing, by leveraging intermediate representations that preserve both numerical and contextual fidelity.





%% file: tables/2-real-se.tex
\begin{table*}[tb!]
\centering
\footnotesize
\caption{Chart Information Extraction (CIE) results evaluated by our proposed SCRM metric under: 1) single-set that the model is trained and evaluated on the same dataset, 2) trained on the merging-set that training samples are merged from the real datasets, and 3) trained on the merging-set that training samples are merged from both real and simulated datasets. Note that `Real Merging' is all real data (ChartQA+PlotQA+Chart2Text), `Real\&Sim Merging' refers to real \& simulated data (ChartQA+PlotQA+Chart2Text+SimChart9K), and `C.D.' denotes Closed-source Dataset.}
\scalebox{0.95}{
\renewcommand{\arraystretch}{1}
\begin{tabular}{ccccrcccccccccc}
\toprule

&\multirow{2}*{Val Set} &\multirow{2}*{Model}  & \multirow{2}*{Train Set} 
& $IoU_{thr}$$\rightarrow$ &  &\multicolumn{2}{c} {mPrecision} & & \multicolumn{4}{c} {Precision} & \\ \cmidrule{7-8} \cmidrule{10-14}
& & & &Tolerance $\downarrow$ & & \multicolumn{2}{c} {0.5:0.05:0.95} &  &0.5& 0.75 & 0.95 & 1 (EM) \\
\midrule

& \multirow{15}{*}{\rotatebox[origin=r]{90}{\multirow{2}*{ChartQA}}}
&\multirow{3}*{Matcha~\cite{Liu2022MatChaEV}} & ChartQA+ & strict \ \textcolor{red!60}{\rule{0.5em}{0.5em}} & &\multicolumn{2}{c}{0.5160} & &0.5814 &0.5114 &0.4678 &0.4460 &\\
& & & PlotQA+ &slight\ \textcolor{yellow!60}{\rule{0.5em}{0.5em}} & & \multicolumn{2}{c}{0.6598} & &0.7045&0.6572 &0.6250 &- &\\
& & & C.D. &high \ \textcolor{green!60}{\rule{0.5em}{0.5em}} & &  \multicolumn{2}{c}{0.7161} & &0.7519 &0.7150 &0.6894 &- &\\ \cmidrule{3-14}

& &\multirow{3}*{Deplot~\cite{Liu2022DePlotOV}} & ChartQA+ & strict \ \textcolor{red!60}{\rule{0.5em}{0.5em}} & &\multicolumn{2}{c}{0.6331} & &0.7008 &0.6326 &0.5814 &0.5663 &\\
& & & PlotQA+ &slight\ \textcolor{yellow!60}{\rule{0.5em}{0.5em}} & & \multicolumn{2}{c}{0.7666} & &0.8229&0.7661 &0.7282 &- &\\
& & & C.D. &high \ \textcolor{green!60}{\rule{0.5em}{0.5em}} & &  \multicolumn{2}{c}{0.8150} & &0.8759 &0.8087 &0.7812 &- &\\ \cmidrule{3-14}

& &\multirow{3}*{Our StructChart} &\multirow{3}*{ChartQA} & strict \ \textcolor{red!60}{\rule{0.5em}{0.5em}} & &\multicolumn{2}{c}{0.6770} & &0.7273 &0.6714 &0.6458 &0.6326 &\\
& & & &slight\ \textcolor{yellow!60}{\rule{0.5em}{0.5em}} & & \multicolumn{2}{c}{0.7792} & &0.8220&0.7746 &0.7519 &- &\\
& & & &high \ \textcolor{green!60}{\rule{0.5em}{0.5em}} & &  \multicolumn{2}{c}{0.8274} & &0.8703 &0.8210 &0.8011 &- &\\ \cmidrule{3-14}

& &\multirow{3}*{Our StructChart} &\multirow{3}*{Real Merging} & strict \ \textcolor{red!60}{\rule{0.5em}{0.5em}} & &\multicolumn{2}{c}{0.7017} &&0.7547 &0.6998 &0.6610 &0.6506 &\\
& & & &slight\ \textcolor{yellow!60}{\rule{0.5em}{0.5em}} & & \multicolumn{2}{c}{0.8227} &&0.8674 &0.8201 &0.7926 &- &\\
& & & &high \ \textcolor{green!60}{\rule{0.5em}{0.5em}} & &  \multicolumn{2}{c}{\textbf{0.8591}} &&0.8987 &\textbf{0.8551} &\textbf{0.8362} &- &\\
\cmidrule{3-14}

& &\multirow{3}*{Our StructChart} &\multirow{3}*{Real\&Sim Merging} & strict \ \textcolor{red!60}{\rule{0.5em}{0.5em}} & &\multicolumn{2}{c}{0.7187} &&0.7683 &0.7153 &0.6705 &\textbf{0.6642} &\\
& & & &slight\ \textcolor{yellow!60}{\rule{0.5em}{0.5em}} & & \multicolumn{2}{c}{0.8311} &&0.8761 &0.8301 &0.8001 &- &\\
& & & &high \ \textcolor{green!60}{\rule{0.5em}{0.5em}} & &  \multicolumn{2}{c}{0.8568} &&\textbf{0.8990} &0.8542 &0.8358 &- &\\

\midrule

& \multirow{15}{*}{\rotatebox[origin=b]{90}{PlotQA}}
&\multirow{3}*{Matcha~\cite{Liu2022MatChaEV}} & ChartQA+  & strict \ \textcolor{red!60}{\rule{0.5em}{0.5em}} & &\multicolumn{2}{c}{0.0048} & &0.0089 &0.0048 &0.0036 &0.0036 &\\
& & & PlotQA+ &slight\ \textcolor{yellow!60}{\rule{0.5em}{0.5em}} & & \multicolumn{2}{c}{0.0752} & &0.0909&0.0754 &0.0635 &- &\\
& & & C.D. &high \ \textcolor{green!60}{\rule{0.5em}{0.5em}} & &  \multicolumn{2}{c}{0.0823} & &0.1093 &0.0837 &0.0719 &- &\\ \cmidrule{3-14}

& &\multirow{3}*{Deplot~\cite{Liu2022DePlotOV}} & ChartQA+  & strict \ \textcolor{red!60}{\rule{0.5em}{0.5em}} & &\multicolumn{2}{c}{0.0997} & &0.1532 &0.1021 &0.0641 &0.0629 &\\
& & & PlotQA+ &slight\ \textcolor{yellow!60}{\rule{0.5em}{0.5em}} & & \multicolumn{2}{c}{0.6969} & &0.8664&0.7435 &0.5463 &- &\\
& & & C.D. &high \ \textcolor{green!60}{\rule{0.5em}{0.5em}} & &  \multicolumn{2}{c}{0.7471} & &0.9679 &0.8034 &0.5992 &- &\\ \cmidrule{3-14}

& &\multirow{3}*{Our StructChart} &\multirow{3}*{PlotQA} & strict \ \textcolor{red!60}{\rule{0.5em}{0.5em}} & &\multicolumn{2}{c}{0.1995} &&0.2500 &0.1931 &0.1765 &0.1736 &\\
& & & &slight\ \textcolor{yellow!60}{\rule{0.5em}{0.5em}} & & \multicolumn{2}{c}{0.7848} &&0.8519 &0.7784 &0.7405 &- &\\
& & & &high \ \textcolor{green!60}{\rule{0.5em}{0.5em}} & &  \multicolumn{2}{c}{0.8271} &&0.8922 &0.8223 &0.7861 &- &\\  \cmidrule{3-14}

& &\multirow{3}*{Our StructChart} &\multirow{3}*{Real Merging} & strict \ \textcolor{red!60}{\rule{0.5em}{0.5em}} & &\multicolumn{2}{c}{0.4549} &&0.5855 &0.4525 &0.3521 &0.3385 &\\
& & & &slight\ \textcolor{yellow!60}{\rule{0.5em}{0.5em}} & & \multicolumn{2}{c}{0.8589} &&0.9210 &0.8569 &0.8118 &- &\\
& & & &high \ \textcolor{green!60}{\rule{0.5em}{0.5em}} & &  \multicolumn{2}{c}{0.8921} &&0.9466 &0.8860 &0.8557 &- &\\
\cmidrule{3-14}

& &\multirow{3}*{Our StructChart} &\multirow{3}*{Real\&Sim Merging} & strict \ \textcolor{red!60}{\rule{0.5em}{0.5em}} & &\multicolumn{2}{c}{0.4596} &&0.5901 &0.4569 &0.3563 &\textbf{0.3612} &\\
& & & &slight\ \textcolor{yellow!60}{\rule{0.5em}{0.5em}} & & \multicolumn{2}{c}{0.8612} &&0.9234 &0.8590 &0.8138 &- &\\
& & & &high \ \textcolor{green!60}{\rule{0.5em}{0.5em}} & &  \multicolumn{2}{c}{\textbf{0.8998}} &&\textbf{0.9547} &\textbf{0.8935} &\textbf{0.8591} &- &\\

\midrule

& \multirow{9}{*}{\rotatebox[origin=b]{90}{Chart2Text}}
&\multirow{3}*{Our StructChart} &\multirow{3}*{Chart2Text} & strict \ \textcolor{red!60}{\rule{0.5em}{0.5em}} & &\multicolumn{2}{c}{0.1936} &&0.2473 &0.1892 &0.1533 &0.1442 &\\
& & & &slight\ \textcolor{yellow!60}{\rule{0.5em}{0.5em}} & & \multicolumn{2}{c}{0.5524} &&0.6603 &0.5529 &0.4672 &- &\\
& & & &high \ \textcolor{green!60}{\rule{0.5em}{0.5em}} & &  \multicolumn{2}{c}{0.6945} &&0.7676 &0.6934 &0.6356 &- &\\ \cmidrule{3-14}

& &\multirow{3}*{Our StructChart} &\multirow{3}*{Real Merging} & strict \ \textcolor{red!60}{\rule{0.5em}{0.5em}} & &\multicolumn{2}{c}{0.3156} &&0.4002 &0.3123 &0.2509 &0.2318 &\\
& & & &slight\ \textcolor{yellow!60}{\rule{0.5em}{0.5em}} & & \multicolumn{2}{c}{0.7141} &&0.7938 &0.7205 &0.6426 &- &\\
& & & &high \ \textcolor{green!60}{\rule{0.5em}{0.5em}} & &  \multicolumn{2}{c}{0.8085} &&0.8595 &0.8090 &0.7673 &- &\\
\cmidrule{3-14}

& &\multirow{3}*{Our StructChart} &\multirow{3}*{Real\&Sim Merging} & strict \ \textcolor{red!60}{\rule{0.5em}{0.5em}} & &\multicolumn{2}{c}{0.3394} &&0.4261 &0.3367 &0.2749 &\textbf{0.2635} &\\
& & & &slight\ \textcolor{yellow!60}{\rule{0.5em}{0.5em}} & & \multicolumn{2}{c}{0.7759} &&0.8522 &0.7701 &0.7096 &- &\\
& & & &high \ \textcolor{green!60}{\rule{0.5em}{0.5em}} & &  \multicolumn{2}{c}{\textbf{0.8296}} &&\textbf{0.8791} &\textbf{0.8287} &\textbf{0.7700} &- &\\
\bottomrule 
\end{tabular}
}
\label{tab:real}
\end{table*}

%% file: tables/3-sim-se.tex
\begin{table}[!tb]
\centering
\footnotesize
\caption{ ChartQA perception results by scaling up the simulation data (from 0.1K to 9K).}
\scalebox{0.88}{
\renewcommand{\arraystretch}{1.2}
\begin{tabular}{clccc}
\toprule

\qquad \qquad &Train Set & mPrecision  & mPrecision &  mPrecision\\
& & @ strict \ \textcolor{red!60}{\rule{0.5em}{0.5em}}& @ slight \ \textcolor{yellow!60}{\rule{0.5em}{0.5em}} & @ high \ \textcolor{green!60}{\rule{0.5em}{0.5em}}\\
\midrule

& ChartQA (w/o sim data)    & 0.6770 & 0.7792 & 0.8274 \\

\midrule 

 \multirow{5}{*}{\rotatebox[origin=c]{90}{ Real \& Sim}}
 & ChartQA + SimChart0.1K &0.6804 &0.7893 &0.8326 \\
\cmidrule{2-2}

 & ChartQA + SimChart1K   & 0.6871  &0.7938 &0.8394 \\
\cmidrule{2-2}

 & ChartQA + SimChart6K  & \underline{0.7040} &\underline{0.8128}  &\underline{0.8450}\\
\cmidrule{2-2}

 & ChartQA + SimChart9K & \textbf{0.7116} &\textbf{0.8182} &\textbf{0.8527}\\
\bottomrule 
\end{tabular}
}
\label{tab:sim}
\end{table}

%% file: tables/4-fewshot.tex
\begin{figure*}
\begin{minipage}[t]{0.48\textwidth}
		\centering
        \captionof{table}{Zero-shot and few-shot CIE performance on real-world chart dataset by means of the proposed SimChart9K.}
        \vspace{-6pt}
		\label{tab:few-shot}
		\scalebox{0.72}{
            \renewcommand{\arraystretch}{1.2}
            \begin{tabular}{clccc}
            \toprule

            \qquad \qquad &\multirow{2}{*}{Train Set} & mPrecision  & mPrecision &  mPrecision\\
            & & @ strict \ \textcolor{red!60}{\rule{0.5em}{0.5em}}& @ slight \ \textcolor{yellow!60}{\rule{0.5em}{0.5em}} & @ high \ \textcolor{green!60}{\rule{0.5em}{0.5em}}\\
            \midrule
            
             \multirow{5}{*}{\rotatebox[origin=c]{90}{ Real Only}}
            & ChartQA 0.01    & 0.1797 & 0.2384 & 0.2686 \\
            \cmidrule{2-2}
            & ChartQA 0.1    & 0.3616 & 0.4242 & 0.4653 \\
            \cmidrule{2-2}
            & ChartQA 0.5    & 0.5389 & 0.6150 & 0.6603 \\
            \cmidrule{2-2}
            & ChartQA    & \underline{0.6770} & \underline{0.7792} & \underline{0.8274} \\
            
            \midrule 
            
             \multirow{7}{*}{\rotatebox[origin=c]{90}{ Real \& Sim}}
             & SimChart9K &0.0688 &0.1577 &0.2527 \\
            \cmidrule{2-2}
            
             & SimChart9K + ChartQA 0.01    & 0.3074  &0.4672 &0.5402 \\
            \cmidrule{2-2}
            
             & SimChart9K + ChartQA 0.1    & 0.5973  &0.7466 &0.7980 \\
            \cmidrule{2-2}
            
             & SimChart9K + ChartQA 0.2    & 0.6465  &0.7787 &0.8206 \\
            \cmidrule{2-2}
            
             & SimChart9K + ChartQA 0.5    & \textbf{0.6902}  & \textbf{0.8015} &\textbf{0.8380} \\
            \bottomrule 
            \end{tabular}
	}
\end{minipage}
\begin{minipage}[t]{0.48\textwidth}
		\centering
        \captionsetup{skip=15pt}
        
        \captionof{table}{Generalizability study for few-shot CIE performance on PlotQA and Chart2Text by means of SimChart9K.}
		\label{tab:few-shot-rebuttal}
        \vspace{-6pt}
		\scalebox{0.78}{
            \renewcommand{\arraystretch}{1.2}
            \begin{tabular}{clccc}
            \toprule

            \multirow{2}{*}{Val Set} &\multirow{2}{*}{Train Set} & mPrecision  & mPrecision &  mPrecision\\
            & & @ strict \ \textcolor{red!60}{\rule{0.5em}{0.5em}}& @ slight \ \textcolor{yellow!60}{\rule{0.5em}{0.5em}} & @ high \ \textcolor{green!60}{\rule{0.5em}{0.5em}}\\
            \midrule
            
             \multirow{4}{*}{\rotatebox[origin=c]{90}{ PlotQA}}
            & \multirow{2}{*}{PlotQA}    & \multirow{2}{*}{0.1995} & \multirow{2}{*}{0.7848} & \multirow{2}{*}{0.8271} \\
            &  &  &  & \\
            \cmidrule{2-2}
            & SimChart9K & \multirow{2}{*}{0.1887} & \multirow{2}{*}{0.7976} & \multirow{2}{*}{0.8063} \\
            & + PlotQA 0.1 &  &  & \\
            \midrule 
             \multirow{4}{*}{\rotatebox[origin=c]{90}{Chart2Text}}
            & \multirow{2}{*}{Chart2Text}    & \multirow{2}{*}{0.1936} & \multirow{2}{*}{0.5524} & \multirow{2}{*}{0.6945} \\
            &  &  &  & \\
            \cmidrule{2-2}
            & SimChart9K & \multirow{2}{*}{0.2610} & \multirow{2}{*}{0.5711} & \multirow{2}{*}{0.6871} \\
            & + Chart2Text 0.2 &  &  & \\
            \bottomrule 
            \end{tabular}
	}
\end{minipage}
\vspace{5pt}
\end{figure*}

%% file: tables/6-chartqa.tex
\begin{table*}[!tb]
\caption{Question-Answering (QA) results on ChartQA, where 'LCT' and our 'STR' are two different chart data schemas. Relaxed-acc metric is employed for evaluation. ‘GT’ indicats the Ground Truth of chart perception results.}
\vspace{-6pt}
\centering
\footnotesize
\begin{tabular}{llcccccccccccccccc}
\toprule
&\multirow{2}*{Model} &\multirow{2}*{Train Data \ }
& \multicolumn{3}{c}{ChartQA val} & & \multicolumn{3}{c}{ChartQA test} \\  \cmidrule{4-6} \cmidrule{8-10}    
& & & aug. \textcolor{blue!30}{\rule{0.7em}{0.55em}} & human \textcolor{red!30}{\rule{0.7em}{0.55em}} & avg. & & aug. \textcolor{blue!30}{\rule{0.7em}{0.55em}} & human \textcolor{red!30}{\rule{0.7em}{0.55em}} & avg. \\

\midrule
  \multirow{6}{*}{\rotatebox[origin=c]{90}{ \textbf{Baseline} }} 
& VL-T5-OCR~\cite{Masry2022ChartQAAB}  &ChartQA & - & - & 41.6 & & - & - & -&\\
& Tapas-OCR~\cite{Masry2022ChartQAAB} &ChartQA& - & - & 45.5 & & - & - & -&\\ 
& PaLI-17B~\cite{Chen2022PaLIAJ} &ChartQA& 64.9 & 30.4 & 47.6 & & - & - & - &\\ 
& Pix2Struct~\cite{Lee2022Pix2StructSP} &ChartQA & 81.6 & 30.5 & 56.0 & & - & - & - & \\ 
& MatCha~\cite{Liu2022MatChaEV} & ChartQA/PlotQA/C.D. & 90.2  & 38.2 & 64.2 & & - & - & - & \\
& Deplot~\cite{Liu2022DePlotOV} & ChartQA/PlotQA/C.D. & 69.3  & 36.6 & 52.9 & & - & - & - & \\
&ChartLlama~\cite{Han2023ChartLlamaAM} &C.D. &90.4 &49.0 &69.7 & &93.1 &58.4 &75.7 \\
&GPT-4V~\cite{openai2023gpt4v} &C.D. &76.1 &64.5 &70.3 & &80.8 &61.4 &75.5 \\
\midrule 
 \multirow{6}{*}{\rotatebox[origin=c]{90}{\textbf{LCT}}}
 &StructChart+Llama2-13b &ChartQA+SimChart9K &61.2 &29.5 &45.4 & &70.1 &27.9 &49.0 &   \\
 &StructChart+Vicuna-13b &ChartQA+SimChart9K &64.6 &28.4 &46.5 & &74.1 &26.2 &50.2 & \\ 
 &StructChart+Llama3.1-8b &ChartQA+SimChart9K &69.8 &41.7 &55.8 & &74.0 &37.4 &55.7 & \\
 &StructChart+Llama3.1-70b &ChartQA+SimChart9K &83.2 &58.9 &71.1 & &90.8 &57.9 &74.3& \\
 &StructChart+GPT-4 &ChartQA+SimChart9K &84.1 &63.1 &73.6 & &91.9 &61.8 &76.8\\
 &GT+GPT-4 &- &87.9 &64.1 &76.0 & &93.8 &63.3 &79.0\\
\midrule
\multirow{6}{*}{\rotatebox[origin=c]{90}{\textbf{STR}}}
 &StructChart+Llama2-13b &ChartQA+SimChart9K &71.2 &33.2 &52.2 & &76.9 &31.7 &54.3    \\
 &StructChart+Vicuna-13b &ChartQA+SimChart9K &79.7 &34.4 &57.1 &  &85.7 &32.8 &59.3   \\ 
 &StructChart+Llama3.1-8b &ChartQA+SimChart9K &72.7 &55.3 &64.0 & &78.3 &54.4 &66.4 \\
 &StructChart+Llama3.1-70b &ChartQA+SimChart9K &84.9 &60.6 &72.8 & &92.2 &59.3 &75.4 \\
 &StructChart+GPT-4 &ChartQA+SimChart9K &85.6 &64.5 &75.5 & &93.5 &63.5 &78.5\\
 &GT+GPT-4 &- &89.7 &65.9 &77.8 & &95.0 &64.2 &79.6\\
\bottomrule 
\end{tabular}

\label{tab:qa}
\end{table*}

%% file: section/5-conclusion.tex
\section{Conclusion and Outlook}
\label{sec:conclusion}

This work has addressed the task of extracting and understanding the structured information from a visual chart. To enhance objectivity and precision in chart related tasks, we introduce \textbf{StructChart}, a novel framework involving the \textbf{S}tructured \textbf{T}riplet \textbf{R}epresentations (STR) for plot-to-triplet chart data shema, the \textbf{S}tructuring \textbf{C}hart-oriented \textbf{R}epresentation \textbf{M}etric (SCRM) for evaluating chart perception tasks, and synthetic dataset \textbf{SimChart9K} for augmentation. Besides, this work has verified that leveraging LLMs to generate more query data and drawing codes, combined with our proposed STR for structured chart descriptions, can enhance the generalization ability of existing large models in understanding chart-related tasks, \textit{e.g.} few-shot chart perception, chart redrawing, and question answering.

For future work, our aim is to further explore the statistical patterns embedded in given chart data to improve the accuracy of time-series forecasting. Integrating multimodal information (\textit{e.g.}, text, tables) with data from the chart domain and leveraging domain knowledge can further enhance predictive performance. Besides, developing a robust, real-time framework that unifies chart understanding, time-series modeling, and cross-modal reasoning remains a key challenge and promising direction.
